\documentclass{article}

\usepackage[final,nonatbib]{nips_2017}

\usepackage[utf8]{inputenc}
\usepackage[T1]{fontenc}
\usepackage{hyperref}
\usepackage{url}
\usepackage{booktabs}
\usepackage{amsfonts}
\usepackage{nicefrac}
\usepackage{microtype}

\usepackage[pdftex]{graphicx}
\usepackage{amsmath}
\usepackage{bm}
\newcommand{\argmax}{\operatornamewithlimits{argmax}}

\title{Competitive Learning \\ Enriches Learning Representation \\ 
and Accelerates the Fine-tuning of CNNs}

\author{
  Takashi Shinozaki \\
  CiNet, National Institute of Information and Communications Technology\\
  1-4 Yamadaoka, Suita, Osaka 565-0871, Japan \\
  \texttt{tshino@nict.go.jp} \\
}

\begin{document}

\maketitle

\begin{abstract}
In this study, we propose the integration of competitive learning into convolutional neural networks (CNNs) 
to improve the representation learning and efficiency of fine-tuning. 
Conventional CNNs use back propagation learning, and
it enables powerful representation learning by a discrimination task. 
However, it requires huge amount of labeled data,
and acquisition of labeled data is much harder than that of unlabeled data.
Thus, efficient use of unlabeled data is getting crucial for DNNs.
%
To address the problem, 
we introduce unsupervised competitive learning into the convolutional layer,  
and utilize unlabeled data for effective representation learning.
The results of validation experiments using a toy model demonstrated that 
strong representation learning effectively extracted bases of images 
into convolutional filters using unlabeled data, 
and accelerated the speed of the fine-tuning 
of subsequent supervised back propagation learning. 
The leverage was more apparent when the number of filters was sufficiently large, 
and, in such a case, the error rate steeply decreased in the initial phase 
of fine-tuning. 
Thus, the proposed method enlarged the number of filters in CNNs, 
and enabled a more detailed and generalized representation. 
It could provide a possibility of not only deep but broad neural networks.  
\end{abstract}

\section{Introduction}
Deep learning and deep neural networks (DNNs)
are the machine learning methods that use 
a neural network (NN) with many layers, which mimics 
information processing in the human brain. 
They have been used to innovate many types of applications in the real world, such as 
image recognition \cite{Krizhevsky2012},   
audio analysis \cite{Dahl2012,Aytar2016},
and even more artistic works \cite{Gatys2015},
One of the most important keys to the substantial advance of DNNs 
is representation learning
and its result, learning representation. 
Most recent DNNs have been based on 
convolutional NNs (CNNs) \cite{LeCun1989},
and trained using back propagation (BP) learning. 
It enables the representation learning by a discrimination task. 
However, it requires huge amount of labeled data,
and the total amount of labeled data is very limited
compared to those of unlabeled.
Thus, efficient use of unlabeled data is getting crucial for DNNs.

The BP learning basically extract optimal learning representation
for the target discrimination task.
It means the obtained learning representation is not completely generalized,
but has some tendency caused by the task.
In order to improve the generalization,
the representation learning should not depend on any type of tasks,
in other words, any kind of labels.
Therefore, 
a method of representation learning by unlabeled data for CNNs is required,
also from the viewpoint of generalization.

One traditional learning method for NN, 
called competitive learning \cite{Rumelhart1985},
is well-known for its strong representation learning. 
It is mainly applied for a self-organizing map, 
which is a single layer NN with a cortex-like topological structure
\cite{Kohonen1982}, and Neocognitron, which is the progenitor 
of recent CNNs \cite{Fukushima1980}. 
Competitive learning maximally uses input dataset information, 
and robustly obtains a large amount of filters with a high degree of freedom.
However, it is not a supervised learning method and 
could not be applied for fine-tuning in the same manner
as supervised BP learning.
Therefore, if competitive learning can unify BP learning on CNNs, 
it is expected that it will enable more detailed
and generalized recognition by DNNs. 

In this study, we propose a method to both integrate competitive learning 
into CNNs and enable effective pre-training using 
unlabeled data. 
We validated the method using a toy model on the MNIST handwritten dataset, 
and demonstrated that competitive learning accelerated
the subsequent supervised BP learning and reduced the number of samples 
of labeled data in the initial learning stage.

\section{Related Work}
There are a vast number of previous studies on the
unsupervised learning of visual representation. 
It was initially studied using analytical methods.
Some previous studies applied independent component analysis to obtain
visual bases from natural scene images \cite{Olshausen1997,Hyvarinen2001}.
Then, the method expanded to a NN-like regime \cite{Le2012}.
Because these methods were too different from conventional NNs
that were trained using BP learning, the unification between neural and
non-neural mechanisms has had some difficulties.

By contrast, as a NN-like architecture, the
Boltzmann machine and its families are applied
for pre-training to initialize the weight parameters
\cite{Hinton2002,Hinton2006}. 
However, the structure and dynamics of these two NNs are
completely different. Thus, there is some inefficiency that results from converting
from the obtained learning representation in the Boltzmann machine
to a target feedforward NN.

Autoencoder \cite{Bengio2007} might address the conversion problem.
It is a variation of a feedforward NN, and
the weight parameter is completely applicable to a traditional feedforward NN.
However, 
CNNs require little more additional processes 
because of the non-linearity of deconvolution processes.
Thus, generator-based unsupervised learning is used for the smooth
transfer of representation learning \cite{Radford2016}.
Goroshin applied a sophisticated approach that extracts
visual bases directly into the subnetwork of the target CNN
using static object information in movies \cite{Goroshin2015}. 
Our approach also deploys bases directly into a part of the CNN,
but extracts bases using a competitive learning method.

\section{Competitive Learning Integration into CNN}
The proposed method adds the capability of competitive learning 
to a convolutional layer in CNN. 
We use the simplest competitive learning method, that is, 
the winner-take-all (WTA) algorithm, 
and do not use any position information over the filter axis. 
WTA is only applied to the process of the weight parameter update 
and has no effect on the feedforward signal. 
The WTA is performed over the input vector of the $l$-th layer, 
which is described as $\bm{u}_{l}=\bm{W}_{l} \bm{z}_{l-1}$, 
where $\bm{z}_{l-1}$ is the output vector of the previous layer 
and $\bm{W}_{l}$ is the connection matrix. 
If the activation function is a monotone increasing function, 
the unit with the maximum input value is that with the maximum output value. 
We call it the `winner' unit and perform a weight update 
of competitive learning only for that unit. 
Therefore, the weight gradient of the $i$-th unit of the $l$-th layer is 
described as follows: 
\begin{equation}
  \Delta \bm{w}_{l,i}=
  \begin{cases}
    - \rho \bm{z}_{l-1}, & \text{if } i=\argmax_{k} u_{l,k} \\
    0, & \text{otherwise},
  \end{cases}
\end{equation}
where $\rho$ is the learning coefficient of competitive learning, 
and we set it to $0.5 \times 10^{-3}$. 
The weight vector is normalized by L2-norm at every update 
to stabilize competitive learning \cite{Kohonen1982}: 
\begin{eqnarray}
\bm{w'}_{l,i} = (\bm{w}_{l,i} + \Delta \bm{w}_{l,i})
/ |\bm{w}_{l,i} + \Delta \bm{w}_{l,i}|.
\label{eq:norm}
\end{eqnarray}

Competitive learning is performed in some selected convolutional layers, 
typically, the earliest layer. 
The layer performs as a conventional convolutional layer 
for the inference and BP learning processes, 
except that the weight is always normalized. 

Competitive learning is treated as unsupervised pre-training. 
First, it extracts the basis 
(e.g., Gabor patches for natural scene images and harmonics for audio)
from unlabeled input data. 
Then, BP learning is applied for fine-tuning, 
in the same manner as conventional CNNs, using learning representation 
obtained by competitive learning. 

We implemented the proposed method with GPU support 
in Python and a Chainer deep learning framework \cite{Tokui2015}. 
We will upload the code of the proposed method 
in the near future.

\section{Experiments}

\begin{figure}[tb]
  \begin{center}
  \raisebox{0.22\linewidth}{\bf \large (a)}\hspace{-0.6ex}
  \includegraphics[width=0.40\linewidth]{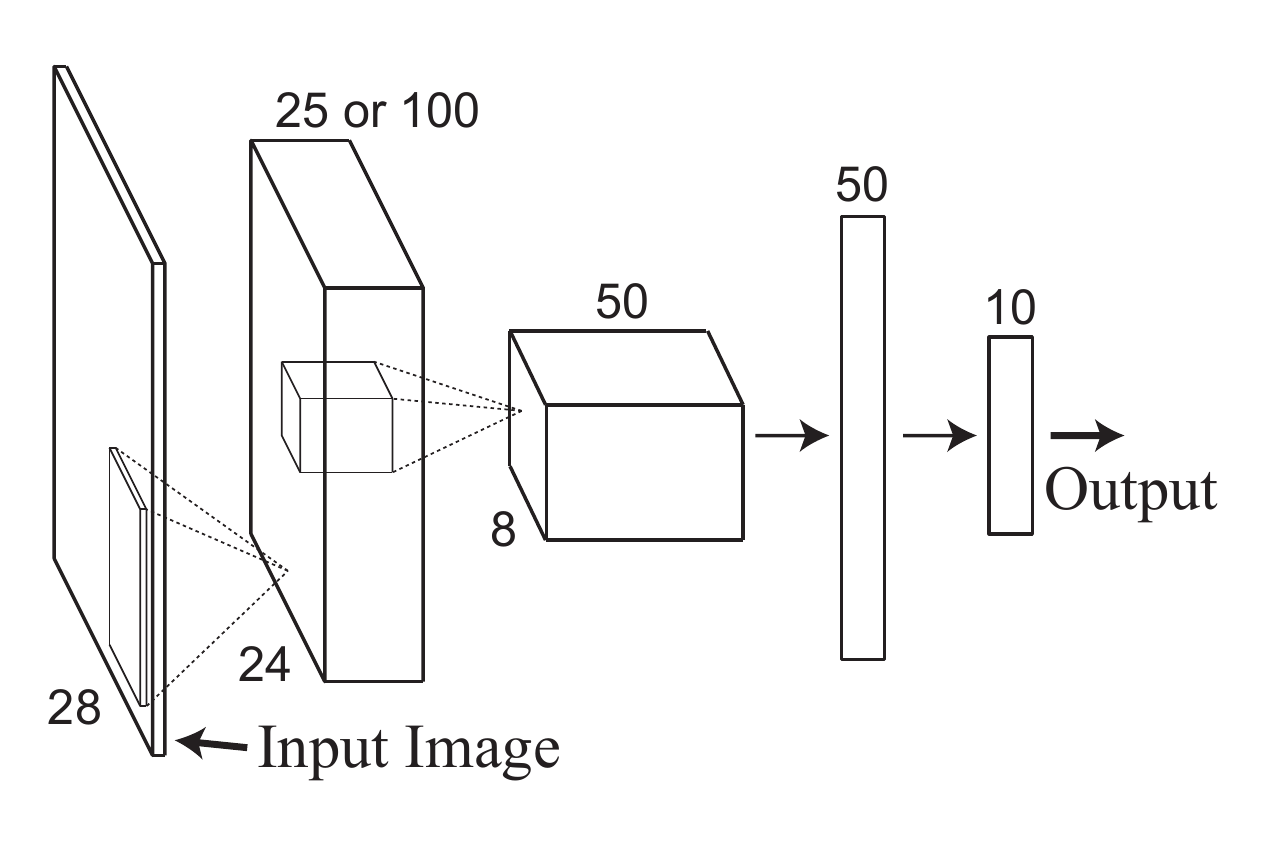}
  \raisebox{0.22\linewidth}{\bf \large (b)}\hspace{-0.6ex}
  \raisebox{0.02\linewidth}{
    \includegraphics[width=0.225\linewidth]{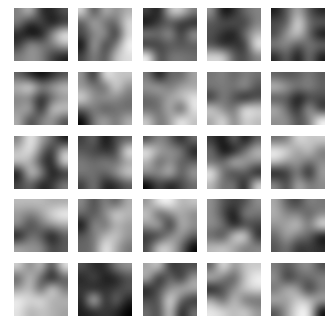}
  }
  \raisebox{0.22\linewidth}{\bf \large (c)}\hspace{-0.6ex}
  \raisebox{0.02\linewidth}{
    \includegraphics[width=0.225\linewidth]{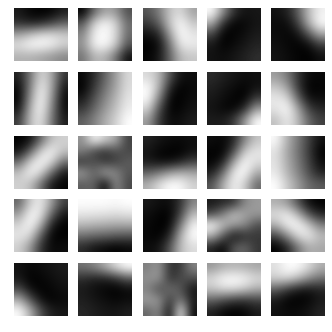}
  }
  \caption{
    (a) Network structure of a simple CNN for our toy model 
    composed of two convolutional layers and two fully connected layers; 
    we integrate competitive learning only for convolutional layers. 
    Obtained learning representations: 
    (b) BP learning of conventional CNN (control condition). 
    (c) Competitive learning. 
  }
  \label{fig:struct}
  \end{center}
\end{figure}

To validate the proposed method, 
we performed an image recognition task on the  
MNIST handwritten dataset \cite{LeCun1998} using a toy model.
Figure \ref{fig:struct}(a) shows the network structure of the model. 
It was composed of two convolutional layers 
with ReLU activation \cite{Nair2010} and max-pooling, 
followed by two fully connected layers. 
The resolution of filters of the convolutional layers were 
5 x 5 and 2 x 2 pixels, 
and the numbers of filters were 25 and 50, respectively. 
The number of units for four layers were 24 x 24, 8 x 8, 100, and 10. 
The input was images with 28 x 28 pixel resolution, and 
the output was 10 categories. 
All weight parameters were initialized to normal random numbers. 

Both pre-training and fine-tuning used 
a training dataset containing 60,000 samples
and the validation process used a test dataset containing 10,000 samples.
The size of the mini-batch was 100 and averaged cross entropy 
was used for the loss function. 
Both competitive learning and BP learning used the
normal stochastic gradient descent method \cite{Amari1967}
for the weight update, and the learning efficient was 0.01. 
We did not use either weight decay or momentum. 
The number of iterations for 
pre-training using competitive learning 
and fine-tuning using BP learning were 15,000 and 5,000, respectively.

Figure \ref{fig:struct}(b) and (c) shows filters 
in the first layer for each learning condition. 
The conventional CNN performed relatively weak representation learning 
and could not obtain a clear spatial pattern on the filters
(Figure \ref{fig:struct}(b)), 
despite the low discrimination error (< 0.02\%). 
By contrast, competitive learning strongly extracted the bases of images
into the filters (Figure \ref{fig:struct}(c)). 
The filters were robust and preserved after fine-tuning 
using BP learning. 

\begin{figure}[tb]
  \begin{center}
  \raisebox{0.35\linewidth}{\bf \large (a)}\hspace{-1.0ex}
  \raisebox{0.03\linewidth}{
    \includegraphics[width=0.45\linewidth]{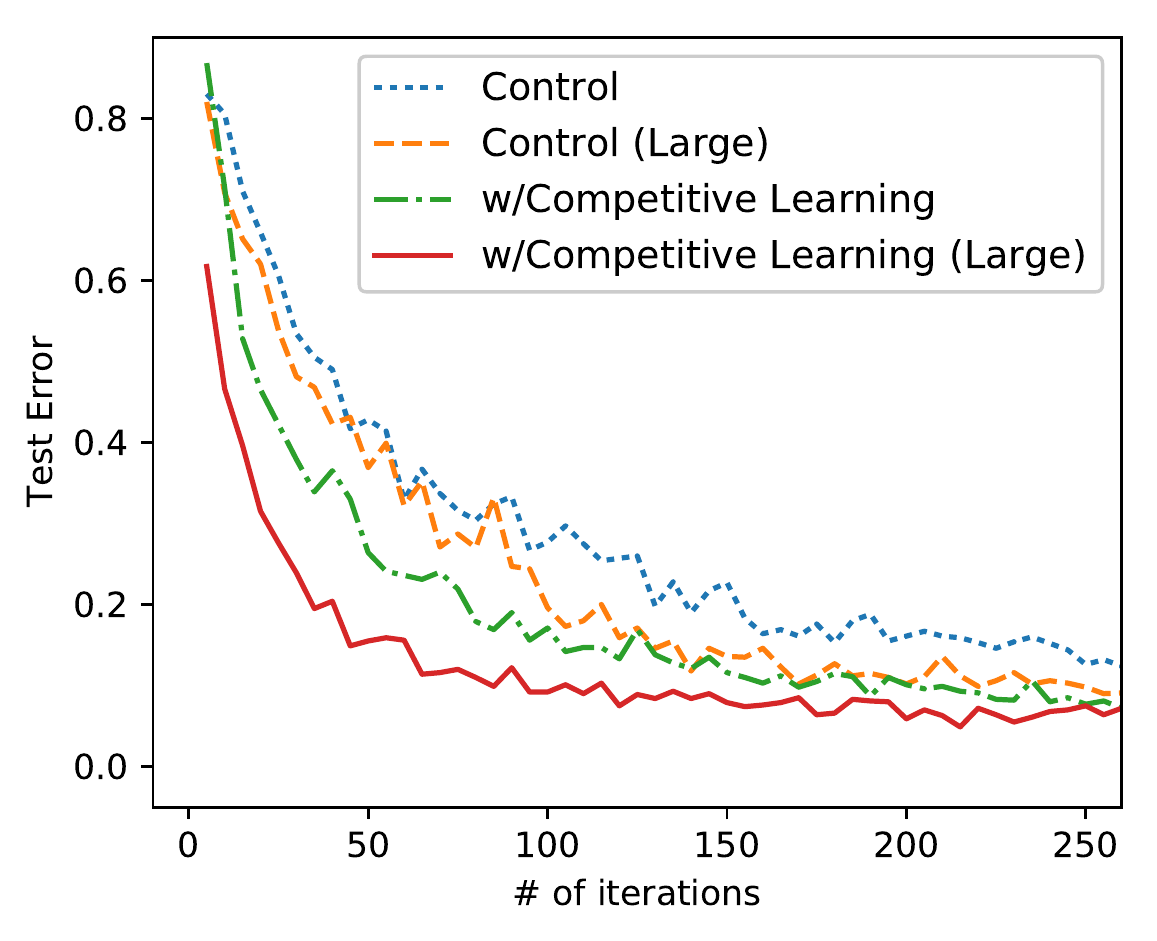}
  }
  \hspace{0.5ex}
  \raisebox{0.41\linewidth}{\bf \large (b)}\hspace{-0.5ex}
  \includegraphics[width=0.45\linewidth]{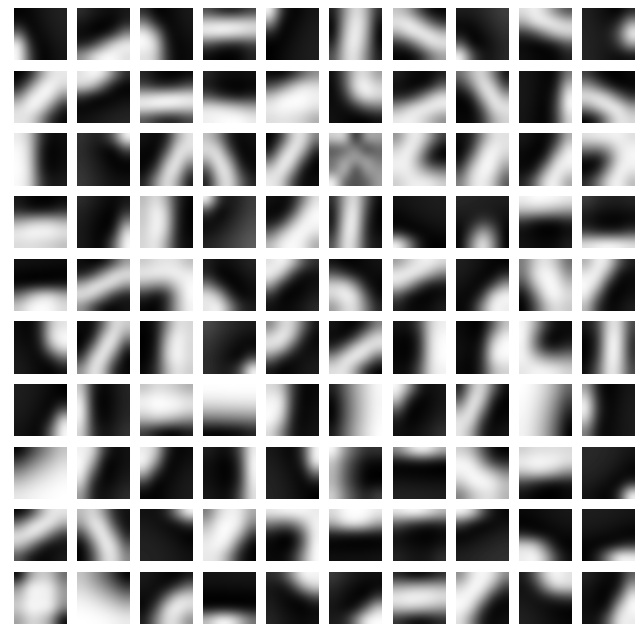}
  \caption{
    (a) Transitions of test errors during fine-tuning. 
    (b) Learning representations for a large number of filters
    in the first convolutional layer. 
  }
  \label{fig:learning}
  \end{center}
\end{figure}

Figure \ref{fig:learning}(a) represents the transition of test errors 
during fine-tuning. 
The horizontal axis represents the number of iterations for supervised BP learning, 
and does not include those of competitive learning. 
Pre-training using competitive learning 
strongly accelerated the fine-tuning process 
because of a better learning representation, in particular, in the initial part. 
The error rates after 5,000 iterations of fine-tuning were 
2.2\% and 1.1\% for CNNs without and with competitive learning, respectively. 
This demonstrates that 
better learning representation obtained better discrimination performance.

Because competitive learning was usable for the usual number of filters, 
it might be suitable for a large number of filters. 
To validate the power of many more filters, 
we increased the number of filters in the first layer 
from 25 to 100 and tested the performance. 
The number of iterations for pre-training using competitive learning was set to 30,000 
because of the enlarged size of the network.
Figure \ref{fig:learning}(b) shows that the obtained filters 
in the first convolutional layer were more detailed and diverse. 
With the rich learning representation, 
the discrimination performance greatly improved, 
as shown in Figure \ref{fig:learning}(a). 
The error rates after 5,000 iterations of fine-tuning were 
1.0\% and  0.9\% for CNNs without and with competitive learning, respectively. 
Although the final error rates were not markedly different between the
two conditions, the steepness in the initial learning phase was 
completely different. 
Thus, quick initial learning is more important 
for a real-world application with a small amount of labeled learning data.
The proposed method can compensate for its performance 
using information from unlabeled data. 
Overall, the results suggests that 
competitive learning can obtain a valuable learning representation 
from unlabeled data, and use them to 
accelerate the fine-tuning process.

\section{Conclusion}
In this study, we proposed the integration of competitive learning into CNN 
to utilize unlabeled data for effective representation learning.
We introduced the WTA mechanism into the convolutional layer 
and utilized unlabeled data and its feedforward information,
enabling a detailed and generalized representation
without any label bias.
The results of the validation experiments using a toy model demonstrated that 
the strong representation learning by the proposed method 
effectively extracted bases of images into filters 
of the convolutional layer and accelerated the speed of fine-tuning 
subsequent BP learning. 
The leverage was more apparent when the number of filters was sufficiently large, 
and, in such a case, the error rate steeply decreased during the initial phase 
of fine-tuning. 

An advantage of the proposed method is stronger representation learning 
than that provided by conventional BP learning. 
Stronger representation learning means stronger adaptability 
for many types of data. 
Moreover, spatial continuity in the input data is not assumed. 
Many recent CNNs have used a spatially small filter size because of 
the relative weakness of representation learning. 
A small filter size requires a large number of layers, 
and it is not good for converging two sparse signals that are spatially divided. 
The strong representation learning of the proposed method 
could converge spatially spread information using a sufficient number of large filters, 
and could also be valuable for high-resolution tasks. 
Moreover, the proposed method is also good for increasing the number of filters. 
Conventional CNNs have weaker representation learning in early layers 
and might have difficulty in increasing the number of filters 
near the input side because of the degradation of the BP signal.
The proposed method could enlarge the number of filters in CNNs, 
and enable more detailed and diverse representation. 

Finally, because the implementation of the proposed method enables 
seamless switching between
unsupervised learning by competitive learning and
supervised learning by BP learning, 
the method could also be useful for the mixture condition of two 
learning methods as semi-supervised learning. 
Further studies are required.

\end{document}